\documentclass[a4paper]{article}

\usepackage[utf8]{inputenc}
\usepackage{erk}
\usepackage{times}
\usepackage{graphicx}
\usepackage[top=22.5mm, bottom=22.5mm, left=22.5mm, right=22.5mm]{geometry}
\usepackage{float}
\usepackage{hyperref}
\hypersetup{
    colorlinks=false,
    }

\usepackage[slovene,english]{babel}

\def\footnotemark{}

\begin{document}
\title{Analysis of face detection, face landmarking, and face recognition performance with masked face images}

\author{Ožbej Golob} 

\affiliation{Faculty of Computer and Information Science,\\
University of Ljubljana, Večna pot 113, 1000 Ljubljana }

\email{E-mail: ozbej.golob@gmail.com}

\maketitle

\begin{abstract}{Abstract}
Face recognition has become an essential task in our lives. However, the current COVID-19 pandemic has led to the widespread use of face masks. The effect of wearing face masks is currently an understudied issue. The aim of this paper is to analyze face detection, face landmarking, and face recognition performance with masked face images. HOG and CNN face detectors are used for face detection in combination with 5-point and 68-point face landmark predictors and VGG16 face recognition model is used for face recognition on masked and unmasked images. We found that the performance of face detection, face landmarking, and face recognition is negatively impacted by face masks.
\end{abstract}

\selectlanguage{english}

\section{Introduction}
Face recognition has become an essential task in our daily life. The wide availability of powerful and low-cost computing systems has popularized face recognition for a variety of applications, including biometric authentication, surveillance, human-computer interaction, and multimedia management \cite{jain2011handbook}. Many highly accurate face recognition models were developed. Some of those models are FaceNet \cite{schroff2015facenet}, DeepFace \cite{Taigman_2014_CVPR}, and DeepID3 \cite{sun2015deepid3}. However, face occlusion can negatively impact the accuracy of face recognition models. This problem has already been addressed in the scope of face detection by Opitz et al. \cite{opitz2016grid}. Some research has also been carried out by Song et al. \cite{Song_2019_ICCV} to develop occlusion invariant face recognition solutions. However, this research focuses on typical in-the-wild occlusion scenarios such as wearing sunglasses.

The current COVID-19 pandemic has led to the wide\-spread use of face masks to prevent the spread of the disease. For this reason, it is essential to research the specific effect of wearing face masks on the performance of face recognition models.

Face landmarking represents one of the first steps in a standard face recognition pipeline. During landmarking, the location of certain facial features (eye corners, the tip of the nose, etc.) is identified in the face images and used to align faces prior to feature extraction. If the landmarks are not detected properly, the alignment procedure will fail and result in poorly aligned or partial facial areas that will ultimately affect face recognition performance. 
In this paper, we explore how face detection, face landmarking, and face recognition work with masked faces.

\section{Related work}
Damer et al. \cite{damer2020effect} study the effect of masked faces on the behavior of three top-performing face recognition systems. Two of these algorithms are academic approaches, namely the ArcFace \cite{deng2019arcface} and SphereFace \cite{liu2017sphereface}. The third algorithm is a commercial off-the-shelf (COTS) from the vendor Neurotechnology \cite{neurotech}. Authors evaluate the face verification performance without masks and compare results to the face verification performance with masks. The verification performance of the ArcFace and SphereFace is negatively affected when the faces are masked, while the COTS is not significantly affected by masked faces.

Wang et al. \cite{wang2020masked} propose three publicly available mask\-ed face datasets: Masked Face Detection Dataset (MFDD), Real-world Masked Face Recognition Dataset (RMFRD), and Simulated Masked Face Recognition Data\-set (SMFRD). Authors propose a face-eye-based multi-gran\-ularity recognition model where they apply different attention weights to key features visible in masked faces (face contour, ocular and periocular details, forehead, etc.). The model improves the recognition accuracy of masked faces from the initial 50\% to 95\%.

\section{Methods}
\subsection{Face detection}
We implemented two face detectors: (i) Histogram of Oriented Gradients (HOG) feature combined with a linear classifier, an image pyramid, and sliding window detection scheme, and (ii) Max-Margin Convolutional Neural Network (CNN) face detector. The HOG detector is accurate and computationally efficient while the CNN detector is accurate and robust, capable of detecting faces from varying viewing angles, lighting conditions, and occlusion. Both face detectors are implemented in Dlib \cite{dlib09}.

\subsection{Face landmarking}
\begin{figure}[!htb]
    \begin{center}
        \includegraphics[width=4cm]{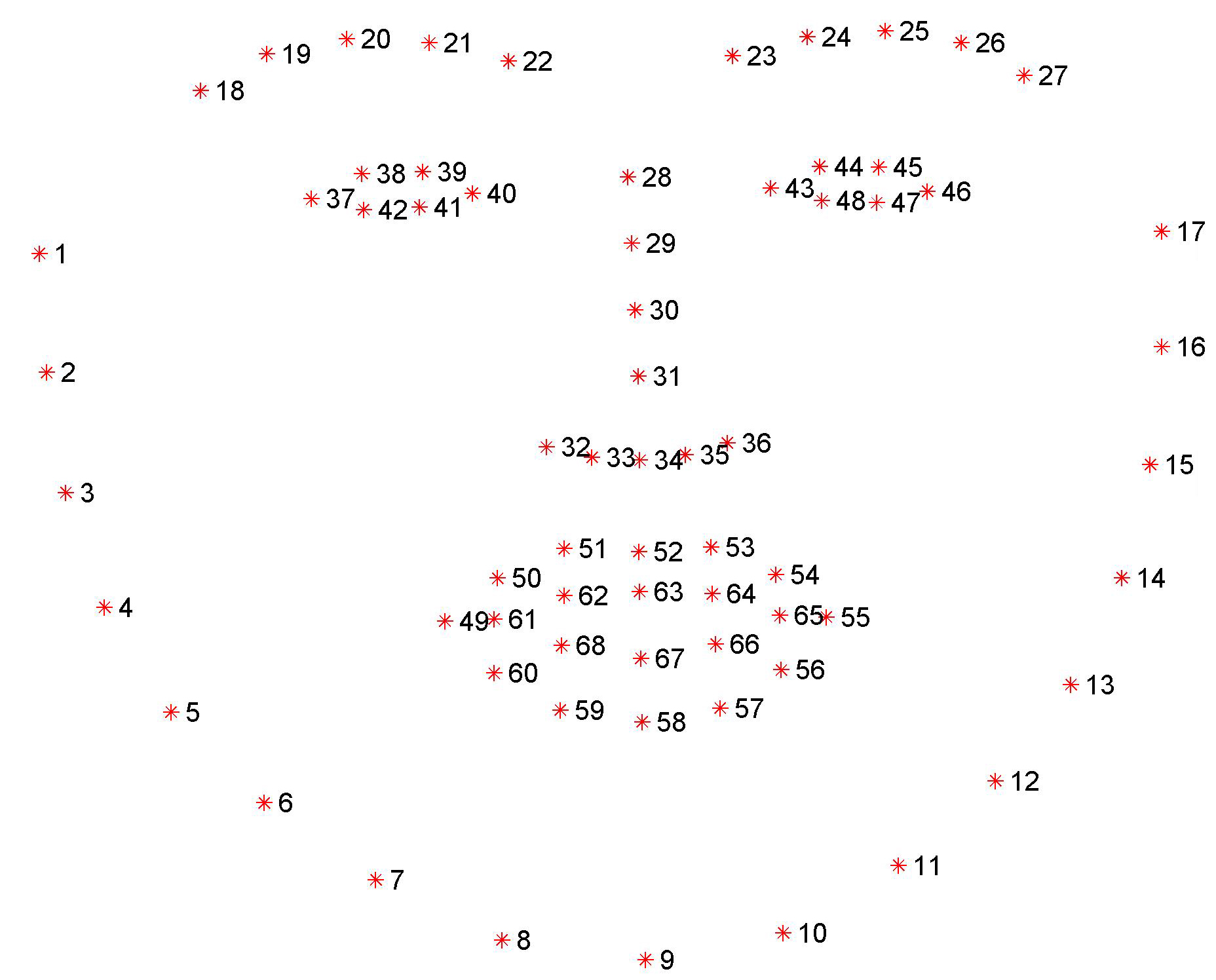}
        \caption{68-point landmarks.} \label{landmarks_68}
    \end{center}
\end{figure}

We implemented two landmark predictors: (i) 68-point landmark predictor (see Figure \ref{landmarks_68}), and (ii) 5-point landmark predictor (see Figure \ref{landmarks_68}, points 34, 37, 40, 43, and 46). Both landmark predictors are Dlib's implementation of \cite{Kazemi_2014_CVPR}.

Face landmarking performance was evaluated with the normalized root mean square error (NRMSE). The normalization is done with respect to the inter-ocular distance (IOD), which is the distance between the two eye centers. Normalizing landmark localization errors with the IOD makes the performance evaluation independent of the face size or the camera zoom factor.

The normalized distance $\delta$ is computed as the Euclidean distance $d(.,.)$ between the ground-truth landmark coordinates $(x,y)$ and the predicted landmark coordinates $(\tilde{x},\tilde{y})$, normalized by the IOD. Equation \ref{landmark_precision_error} shows the formula for the normalized distance of each landmark, where subscript $k$ indicates one of the landmarks.
\begin{equation}
    \label{landmark_precision_error}
    \delta_k = \frac{d\{(x_k,y_k),(\tilde{x_k},\tilde{y_k})\}}{IOD}
\end{equation}

NRMSE of each image ($NRMSE_{local}$) is calculated by the formula shown in Equation \ref{nrmse_local}, where $n$ is the number of landmarks.
\begin{equation}
    \label{nrmse_local}
    NRMSE_{local} = \sqrt{\frac{\displaystyle\sum_{k=1} ^{n}~\delta_k^2}{n}}
\end{equation}

\subsection{Face recognition}
\begin{figure*}[!htb]
    \begin{center}
        \includegraphics[width=15cm]{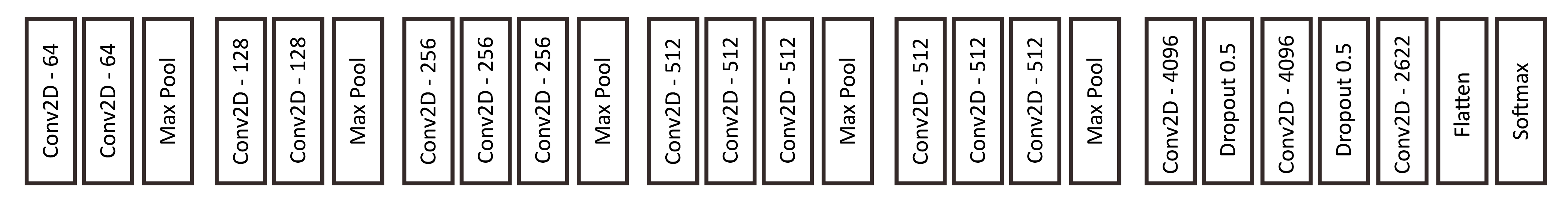}
        \caption{VGG16 architecture.} \label{vgg16}
    \end{center}
\end{figure*}

VGG16 architecture which is pre-trained on a huge ImageNet database with more than 1 million images belonging to 1000 different categories is used to train the input face images. The Softmax layer is removed in order to get image feature vectors. The extracted features are fed as input to the Fully Connected Layer and Softmax activation. Figure \ref{vgg16} shows the VGG16 architecture.

\section{Experiments}
\subsection{Datasets}
To evaluate face landmarking performance, we identified a couple of datasets with annotated facial landmarks. These datasets are Labeled Face Parts in the Wild (LFPW) \cite{belhumeur2013localizing}, Annotated Faces in the Wild (AFW) \cite{zhu2012face}, HELEN \cite{le2012interactive}, and IBUG \cite{sagonas2013300}. All used datasets are collected in the wild. 
The LFPW dataset consists of 1,432 faces, where 1,132 images are a part of the train set and 300 images are a part of the test set.
The HELEN dataset is composed of 2330 face images, where 2000 images are a part of the train set and 330 images are a part of the test set. 
The AFW dataset contains 205 images with 468 faces. 
The IBUG dataset includes 135 images with extreme poses and expressions.
Evaluation on LFPW and HELEN was performed on the test set only while evaluation on AFW and IBUG was performed on the whole dataset.
All used datasets were re-annotated with 68 landmarks as a part of 300 Faces In-the-Wild Challenge (300-W) \cite{300w} using the mark-up of Figure \ref{landmarks_68}.

To evaluate face landmarking performance on masked faces, we generated masked versions of LFPW, AFW, HELEN, and IBUG datasets with the help of the MaskTheFace tool
\cite{anwar2020masked}.

To evaluate face recognition performace, we used a subset of CASIA dataset. 960 train images and 417 test images, belonging to 10 identities, were used for evaluation. To evaluate face recognition performance on masked faces, we generated masked images with the help of the MaskTheFace tool.

\subsection{Experimentation details}
\begin{figure}[H]
    \begin{center}
        \includegraphics[width=4cm]{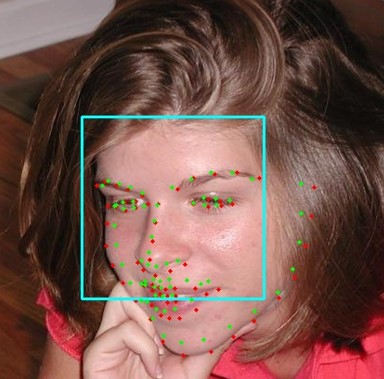}
        \caption{Face landmarks (red dots represent ground truth landmarks, green dots represent predicted landmarks).} \label{landmarks}
    \end{center}
\end{figure}

Face detection results were measured as a percentage of annotated faces that the detector was able to detect. Face landmarking results were measured as NRMSE. Figure \ref{landmarks} shows positions of ground truth and predicted landmarks. The distance between ground truth and predicted landmarks (see Equation \ref{landmark_precision_error}) is calculated for each landmark and NRMSE (see Equation \ref{nrmse_local}) is calculated for each image. NRMSE is then averaged for the whole dataset.

We ignored images where zero faces were detected and images where the annotated face was not detected. Images with errors were ignored because they increased the NRMSE significantly.

VGG16 model for face recognition was initialized with weights learned from a pre-trained model. VGG16 model was additionaly trained on unmasked images and evaluated on unmasked and masked images. Model was then re-trained and evaluated on masked images.

\section{Results}
\subsection{Face detection}
Table \ref{results_face_detection_hog} shows the HOG face detection accuracy on original and masked datasets.
\begin{table}[h]
\caption{HOG face detection accuracy.} \label{results_face_detection_hog}
\smallskip
\begin{center}
\begin{tabular}{ | r | c | c | }
\hline  
  \textbf{Dataset} & \textbf{Original} & \textbf{Masked}\\ 
\hline  
  HELEN & 0.967 & 0.735\\
  LFPW & 0.987 & 0.801\\
  AFW & 0.896 & 0.565\\
  IBUG & 0.711 & 0.487 \\
  \hline  
  Average & 0.890 & 0.647 \\
\hline  
\end{tabular}
\end{center}
\end{table}

Table \ref{results_face_detection_cnn} shows the CNN face detection accuracy on original and masked datasets.
\begin{table}[h]
\caption{CNN face detection accuracy.} \label{results_face_detection_cnn}
\smallskip
\begin{center}
\begin{tabular}{ | r | c | c | }
\hline  
  \textbf{Dataset} & \textbf{Original} & \textbf{Masked}\\ 
\hline  
  HELEN & 1.000 & 0.892\\
  LFPW & 1.000 & 0.878\\
  AFW & 1.000 & 0.708\\
  IBUG & 0.941 & 0.635 \\
  \hline  
  Average & 0.985 & 0.778 \\
\hline  
\end{tabular}
\end{center}
\end{table}

\subsection{Face landmarking}
Table \ref{results_face_landmarking_hog_68} shows the Dlib 68-point face landmarking NRMSE with HOG face detector on original and masked datasets.
\begin{table}[ht]
\caption{Dlib 68-point face landmarking NRMSE with HOG face detector.} \label{results_face_landmarking_hog_68}
\smallskip
\begin{center}
\begin{tabular}{ | r | c | c | }
\hline  
  \textbf{Dataset} & \textbf{Original} & \textbf{Masked}\\ 
\hline  
  HELEN & 0.044 & 0.153\\
  LFPW & 0.054 & 0.159\\
  AFW & 0.063 & 0.169\\
  IBUG & 0.102 & 0.204 \\
  \hline  
  Average & 0.066 & 0.171 \\
\hline  
\end{tabular}
\end{center}
\end{table}

Table \ref{results_face_landmarking_cnn_68} shows the Dlib 68-point face landmarking NRMSE with CNN face detector on original and masked datasets.
\begin{table}[ht!]
\caption{Dlib 68-point face landmarking NRMSE with CNN face detector.} \label{results_face_landmarking_cnn_68}
\smallskip
\begin{center}
\begin{tabular}{ | r | c | c | }
\hline  
  \textbf{Dataset} & \textbf{Original} & \textbf{Masked}\\ 
\hline  
  HELEN & 0.065 & 0.187\\
  LFPW & 0.073 & 0.189\\
  AFW & 0.114 & 0.214\\
  IBUG & 0.209 & 0.266 \\
  \hline  
  Average & 0.115 & 0.214 \\
\hline  
\end{tabular}
\end{center}
\end{table}

Table \ref{results_face_landmarking_hog_5} shows the Dlib 5-point face landmarking \\NRMSE with HOG face detector on original and masked datasets.
\begin{table}[ht]
\caption{Dlib 5-point face landmarking NRMSE with HOG face detector.} \label{results_face_landmarking_hog_5}
\smallskip
\begin{center}
\begin{tabular}{ | r | c | c | }
\hline  
  \textbf{Dataset} & \textbf{Original} & \textbf{Masked}\\ 
\hline  
  HELEN & 0.034 & 0.105\\
  LFPW & 0.050 & 0.113\\
  AFW & 0.062 & 0.123\\
  IBUG & 0.092 & 0.172 \\
  \hline  
  Average & 0.060 &  0.128 \\
\hline  
\end{tabular}
\end{center}
\end{table}

Table \ref{results_face_landmarking_cnn_5} shows the Dlib 5-point face landmarking \\NRMSE with CNN face detector on original and masked datasets.
\begin{table}[!ht]
\caption{Dlib 5-point face landmarking NRMSE with CNN face detector.} \label{results_face_landmarking_cnn_5}
\smallskip
\begin{center}
\begin{tabular}{ | r | c | c | }
\hline  
  \textbf{Dataset} & \textbf{Original} & \textbf{Masked}\\ 
\hline  
  HELEN & 0.037 & 0.114\\
  LFPW & 0.052 & 0.115\\
  AFW & 0.070 & 0.136\\
  IBUG & 0.107 & 0.152 \\
  \hline  
  Average & 0.067 & 0.129 \\
\hline  
\end{tabular}
\end{center}
\end{table}

\subsection{Face recognition}
VGG16 model trained on unmasked images achieved test accuracy of 0.966 on unmasked images and 0.867 on masked images. After the model was re-trained on masked images, the model achieved 0.952 accuracy on masked images.

\section{Conclussion}
In the scope of this paper, we analyzed face detection, face landmarking, and face recognition performance with masked face images.
We observed a negative impact of face masks on face detection. Face detection accuracy dropped by 0.243 on average for the HOG face detector and by 0.207 on average for the CNN face detector for masked images.
Face landmarking performance was also negatively impacted by face masks. NRMSE of the 68-point face landmarking predictor increased nearly 3-times with the HOG face detector and nearly 2-times with the CNN face detector for masked images. NRMSE of the 5-point face landmarking predictor increased by approximately 2-times for HOG and CNN face detector for masked images.
Face recognition performance was also negatively impacted by face masks, lowering the accuracy by 0.099 on masked images. We were able to improve the accuracy to 0.952 by additionally training the model on masked images.

We conclude that face masks have a negative impact on face detection, face landmarking, and face recognition. This implies that facial areas beneath face masks (mouth, nose, etc.) hold significant information for face detection, face landmarking, and face recognition. We found that certain models can be improved by additionally training the model on masked images.

\small
\bibliographystyle{unsrt}
\bibliography{erk}

\end{document}